\documentclass{article}
\PassOptionsToPackage{numbers, compress}{natbib}
\usepackage[preprint]{neurips_2025}

\usepackage[utf8]{inputenc} % allow utf-8 input
\usepackage[T1]{fontenc}    % use 8-bit T1 fonts
\usepackage{hyperref}       % hyperlinks
\usepackage{url}            % simple URL typesetting
\usepackage{booktabs}       % professional-quality tables
\usepackage{amsfonts}       % blackboard math symbols
\usepackage{nicefrac}       % compact symbols for 1/2, etc.
\usepackage{microtype}      % microtypography
\usepackage{xcolor} 
\usepackage{amsmath}% colors
\usepackage{graphicx}
\usepackage{makecell}
\newcommand{\dynaSubVAE}{DynaSubVAE}
\usepackage{algorithm}
\usepackage{algorithmic}
\usepackage{enumitem}
\usepackage{subcaption}
\usepackage{graphicx}

\title{DynaSubVAE: Adaptive Subgrouping for Scalable and Robust OOD Detection}
\author{%
  Tina Behrouzi \\
  Department of Computer Science\\
  University of Toronto and Vector Institute\\
  Toronto, ON, Canada \\
  \texttt{tina.behrouzi@mail.utoronto.ca} \\
  \And
  Sana Tonekaboni \\
  Eric and Wendy Schmidt Center\\
  Broad Institute of MIT and Harvard\\
  Cambridge, MA, USA\\
  \AND
  Rahul G. Krishnan \\
  Department of Computer Science\\
  University of Toronto and Vector Institute\\
  Toronto, ON, Canada \\
  \And
  Anna Goldenberg \\
  Department of Computer Science\\
  Department of Laboratory Pathology and Medicine\\
  University of Toronto and Vector Institute \\
  Toronto, ON, Canada \\
}

\begin{document}

\maketitle

\begin{abstract}
Real-world observational data often contain existing or emerging heterogeneous subpopulations that deviate from global patterns.
% posing challenges for 
% Standardized 
The majority of models tend to overlook these underrepresented groups, leading to inaccurate or even harmful predictions. %Standardized
% Existing Out-of-domain (OOD) detection methods are typically supervised, lack direct subgrouping, or rely on rigid parametric assumptions, often defaulting to abstention in the face of OOD instances rather than adaptation. 
Existing solutions often rely on detecting these samples as Out-of-domain (OOD) rather than adapting the model to new emerging patterns. 
We introduce DynaSubVAE, a Dynamic Subgrouping Variational Autoencoder framework that jointly performs representation learning and adaptive OOD detection. Unlike conventional approaches, DynaSubVAE evolves with the data by dynamically updating its latent structure to capture new trends. It leverages a novel non-parametric clustering mechanism, inspired by Gaussian Mixture Models, to discover and model latent subgroups based on embedding similarity. 
% When encountering distributionally-shifted 
% samples, our framework can spawn new clusters and selectively fine-tune embeddings while preserving global knowledge. 
Extensive experiments show that DynaSubVAE achieves competitive performance in both near-OOD and far-OOD detection, and excels in class-OOD scenarios where an entire class is missing during training. We further illustrate that our dynamic subgrouping mechanism outperforms standalone clustering methods such as GMM and KMeans++ in terms of both OOD accuracy and regret precision. 
% Incorporating streaming OOD samples and retraining the classifier on buffered embeddings further boosts accuracy by over 10\%, demonstrating the framework’s adaptability in dynamic environments.
% Observational data in real-world settings often exhibit high heterogeneity, with subpopulations that deviate significantly from global trends. This poses a critical challenge for standardized models, which frequently overlook emerging or evolving subgroups, leading to poor or even harmful predictions for these underrepresented samples. Existing methods are often supervised, lack direct subgrouping, or rely on parametric assumptions. They typically treat uncertain instances as out-of-distribution (OOD) and default to abstention, rather than enabling model adaptation.
\footnote{Our code will publicly be available on our github.}
\end{abstract}

\section{Introduction}
The safe deployment of machine learning models is essential in high-stakes domains such as healthcare \citep{matos2024clinician} and autonomous driving \citep{yao2024out}, where data heterogeneity poses a major challenge. In observational settings like medical imaging, a new patient’s scan may deviate from population-level patterns due to a novel or rare condition not seen during training. Detecting such out-of-distribution (OOD) samples is crucial to ensure they are handled safely. Yet, traditional models often fail to recognize or adapt to these deviations, leading to poor decisions and unintended consequences \citep{zadorozhny2022out, chen2021probabilistic, sun2023survey}.
While it is often assumed that improving in-distribution (ID) accuracy also enhances OOD generalization \citep{miller2021accuracy}, this assumption breaks down or even reverses in the presence of label noise \citep{sanyal2024accuracy}. This underscores the need for OOD detection methods that remain robust under real-world data imperfections.

A key challenge in OOD detection is identifying OOD samples without relying on explicit labels, which are frequently scarce, missing, or unreliable. However, most existing methods focus on supervised settings \citep{yang2024generalized}. Self-supervised representation learning offers a promising alternative, showing improved robustness to distributional shifts \citep{shirobust}. These learned representations can be leveraged to enhance OOD detection in label-sparse environments \citep{khalid2022rodd, guille2023cadet}. Yet, to remain effective in dynamic real-world scenarios, where data distributions evolve over time, models must move beyond static representation-based OOD detection and learn to adapt continuously. 

% Self-supervised and generative models, which learn representations directly from data without the need for extensive labeled supervision, have emerged as promising alternatives.  Recently, a number of self-supervised approaches have been proposed to enhance OOD detection performance in label-sparse environments \cite{khalid2022rodd, guille2023cadet}. These methods also reduce computational overhead by storing only low-dimensional embeddings, which are then used to measure similarity to ID data. 
% However, to remain effective in dynamic, real-world settings where data distributions evolve over time, OOD detection systems must go beyond static representations. 

% There is a growing need for models that can adapt continuously without relying on ground-truth labels or full retraining. While existing work in online learning [CITE] has explored model updates in supervised settings with known group labels, 

Robust strategies for subgroup discovery in self-supervised contexts, particularly under distributional shift, remain limited \citep{caron2020unsupervised}. Most existing approaches are often restricted to parametric methods with fix structure assumptions that scale poorly and are difficult to apply in real-world scenarios \cite{wani2024comprehensive}. In contrast, streaming clustering approaches, such as DBSTREAM \cite{cao2006density}, offer more flexibility through incremental, batch-driven determination of cluster centers, better suiting evolving or streaming data settings. Recent work on lifelong unsupervised learning using mixup has shown that self-supervised models can offer improved robustness to continual learning challenges, such as catastrophic forgetting, outperforming their supervised counterparts \citep{madaanrepresentational}. 

We introduce \dynaSubVAE, a self-supervised framework for adaptive subgroup detection that enhances OOD detection and representation learning in dynamic environments. \dynaSubVAE\ uniquely integrates OOD detection with dynamic subgrouping, enabling a representation learning model to flexibly encode new and emerging data into evolving latent clusters. Upon detecting an OOD sample as new observations are made, the framework initiates a new cluster and incrementally learns its distribution.
Unlike conventional clustering methods, which require access to the full dataset and assume a fixed number of clusters, \dynaSubVAE\ performs a nonparametric subgrouping using a deep, Gaussian Mixture Model (GMM) \cite{dempster1977maximum} inspired mechanism embedded within a conditional Variational Autoencoder (VAE). Clustering occurs in the latent space, which both reduces computational overhead and improves the model's ability to capture class ood detection. This latent-level analysis is especially useful for distinguishing structural novelty (latent OOD) from surface-level anomalies (sample-level OOD).

\dynaSubVAE\ achieves this by introducing an regret-masking mechanism that constrains model uncertainty on in-distribution data. This sharpens the boundary between familiar and novel distributions, improving sensitivity to gradual or abrupt distributional shifts and supporting robust OOD detection under continuous change. Importantly, \dynaSubVAE\ is designed for scalability; The backbone encoder remains frozen, eliminating the need for full retraining, while a lightweight clustering module maintains and updates statistical structure online. Unlike traditional GMMs, our method avoids storing all original images, making it well-suited for streaming and resource-constrained settings.

Our contributions include: 
\begin{itemize}[leftmargin=*]
    \item We propose \dynaSubVAE, a deep learning-based non-parametric clustering method integrated into the VAE framework, capable of dynamically learning data subgroups.

    \item To the best of our knowledge, this is the first work to quantify 'regret' through uncertainty analysis of subgroup-based interventions within a conditional VAE framework.
    
    \item We incorporate novel clustering loss functions that regulate the creation of new clusters and control weight distribution. This includes an augmentation-based loss that ensures augmented views of the same sample remain within the same subgroup, enhancing robustness for OOD detection. The architecture of the proposed \dynaSubVAE\ introduces a unique incremental embedding mechanism, where each subgroup is associated with specific weights in VAE structure. 
    
    % \item The architecture of the proposed \dynaSubVAE\ introduce a unique incremental embedding, where each subgroup is associated with specific weights, enabling partial weight updates during continual learning without interfering with previously learned representations.
    
    \item We provide a comprehensive evaluation on both simulated and real-world datasets to demonstrate the effectiveness of our subgrouping approach in detecting both near-OOD and far-OOD samples. For example, on the CIFAR-10 dataset, our method reduced FRP@95 by 29\% and improved OOD detection accuracy on SVHN by 21.4\% compared to SOTA models. We also explore its extension to class-based OOD detection, where visual similarity poses greater challenges.
    
\end{itemize}

\section{Related Works}
% In this section, we provide a brief overview of out-of-distribution (OOD) detection methods, as well as deep VAE architectures specifically designed for clustering.
\paragraph{\textbf{Out of Distribution (OOD) Detection: }}
This work \cite{yang2024generalized} presents a comprehensive framework for understanding OOD detection, distinguishing it from related concepts such as anomaly detection (AD), novelty detection (ND), open set recognition (OSR), and outlier detection (OD). We compare our OOD detection method with several state-of-the-art approaches, including KNN \cite{sun2022out}, DICE \cite{sun2022dice}, Fdbd \cite{liu2024fast}, Scale \cite{xu2023scaling}, MSP \cite{hendrycks2016baseline}, as proposed in the paper. A comprehensive analysis for both near and far OOD detection are performed using the logit space to identify OOD data. 
%logitnorm \cite{wei2022mitigating}, 
In these approaches \cite{zhang2023openood}, detecting OOD samples typically requires access to a portion of OOD data to determine an appropriate threshold. 
% Specifically, 20\% of the test set is used to optimize the separation between in-distribution and OOD samples. 
Notably, these methods rely on fully supervised training, assuming the availability of labeled data. %including some OOD examples, to perform threshold calibration.
RODD \cite{khalid2022rodd} is a self-supervised method for detecting out-of-distribution (OOD) samples by computing the k-nearest neighbors of a test sample's embedding with respect to embeddings of in-distribution (ID) training data. However, the method does not support dynamic subgrouping or joint training of clustering and representation learning; the encoder is trained independently, and clustering is applied post hoc during inference. 
% Moreover, the method is not trained to reconstruct images, and therefore cannot be used for image generation based on clusters.
Several existing approaches  \cite{xu2023machine,sun2022out} similarly decouple clustering from representation learning. However, this separation limits their ability to support incremental learning, where new OOD data could form new clusters and only the relevant parts of the network would be updated, without altering the learned representations of the ID data. 

% Provable guarantees for understanding out-of-distribution detection \cite{morteza2022provable}

% \subsection{Dynamic clustering}
% Dynamic Cluster Allocation -> threshold based 

% Reclustering

% Incremental Clustering:

% Online k-mean clustering: calculate the regret based on the clustering performance difference considering current cluster assignment in online setting and the case that all data is available and the cluster is trained offline.

\paragraph{\textbf{Generative Clustering: }}

Accurate subgrouping is essential; for example, in the medical field, it enables the identification of patients with similar patterns, aiding diagnosis and treatment decisions. Variational Deep Embedding (VaDE) \cite{jiang2016variational} encourages the encoder to produce latent representations aligned with Gaussian Mixture Model (GMM) components, thereby implicitly promoting clustering in the latent space. However, VaDE assumes a fixed number of clusters, limiting its capacity for incremental learning or adaptation to new data. Methods such as MCluster-VAE \cite{rong2022mcluster, kingma2014semi} employ a hierarchical structure that combines deep generative models with categorical latent variables, trained in an end-to-end manner. While clustering is learned as part of the variational objective, the number of clusters must be predefined, and the model lacks flexibility to dynamically incorporate new clusters. % or adapt to novel data in an incremental fashion.
Deep Embedded Clustering (DEC) \cite{ren2024deep} adopts a two-stage training process: it first pretrains an autoencoder and then discards the decoder, fine-tuning only the encoder for clustering. However, the selected cluster assignments do not directly influence the learned embeddings. Our approach addresses these limitations by explicitly leveraging subgroup structures to enhance representation learning, which also improves out-of-distribution (OOD) detection. Crucially, our method is non-parametric: the number of clusters is not fixed and can adapt dynamically as new data arrives. This makes it particularly well-suited for streaming or evolving environments, where new classes and patterns may emerge over time.

\section{Methodology}
\begin{figure}[ht]
    \centering
    \includegraphics[width=.9\textwidth]{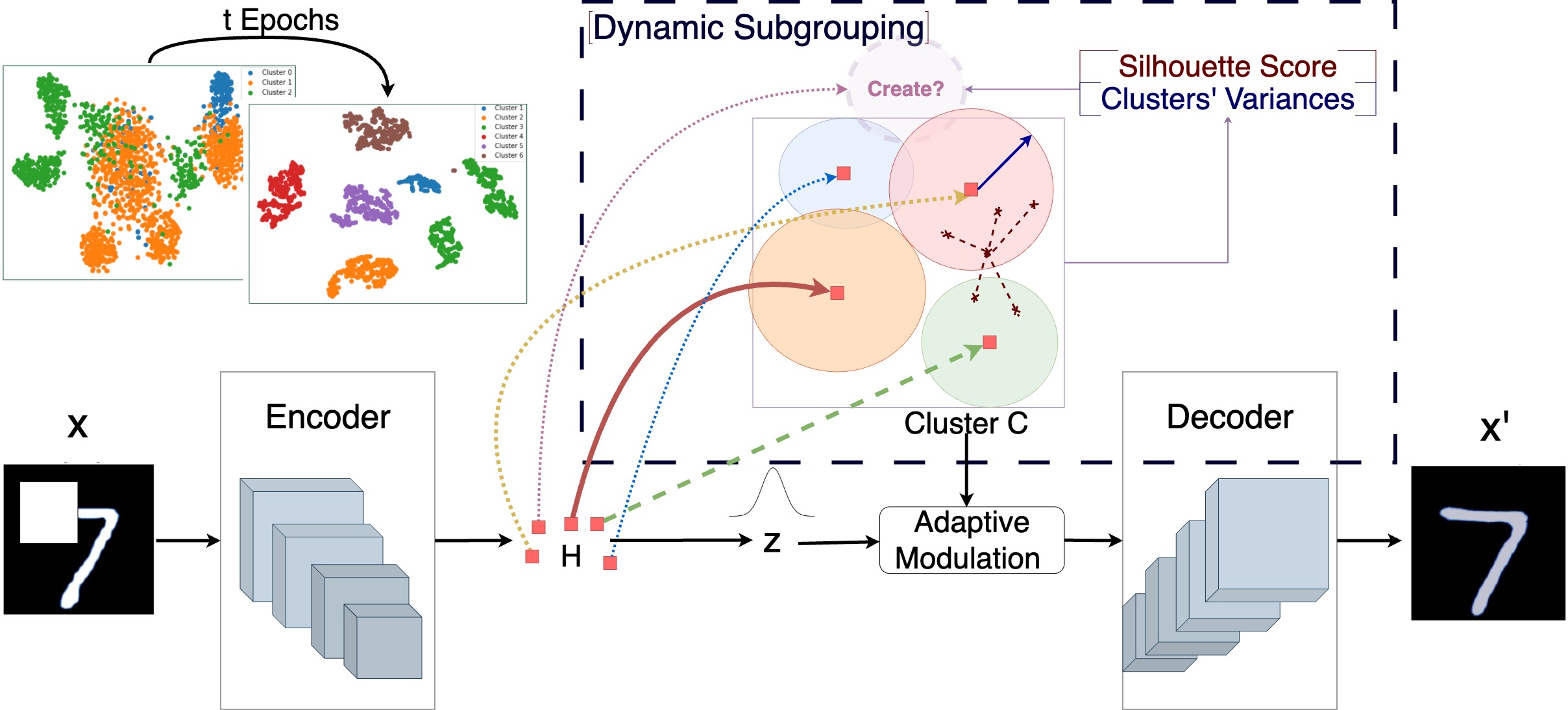}
    \caption{Overview of the \dynaSubVAE\ methodology. The top-left part of the figure illustrates how integrating dynamic subgrouping with the VAE helps push representations further apart, while the subgrouping mechanism increases the number of clusters to better reflect the embeddings' structure.}
    \label{fig:method_overview}
\end{figure}

In this section, we present the detailed structure of \dynaSubVAE. We begin by outlining the OOD detection problem and the overall model workflow, as illustrated in Figure~\ref{fig:method_overview}. We then explain the architecture of the proposed dynamic subgrouping and its associated loss functions in Subsection~\ref{subsection:dynamic_subgrouping}, where we also introduce the regret function based on subgroup assignments. Finally, Subsection~\ref{subsection:VAElosses} describes the representation learning losses that enable the synchronous training of the VAE and subgrouping modules.

\paragraph{Preliminaries:} In the context of the OOD detection problem, we consider a set of in-distribution (ID) data pairs $\mathcal{I}_{\text{ID}} =$ \(\{(x_i^{\text{ID}}, y_i^{\text{ID}})\}_{i=1}^n\), where \(y_i \in \{1, 2, \dots, L\}\) denotes the class labels of ID samples used during training. OOD samples are denoted as \((x_j^{\text{OOD}}, y_j^{\text{OOD}})\), and they differ semantically from the ID data such that \(P(y_j^{\text{OOD}} \mid x_j^{\text{OOD}}) \neq P(y_i^{\text{ID}} \mid x_i^{\text{ID}})\) for any $(x_i^{\text{ID}}, y_i^{\text{ID}}) \sim \mathcal{I}_{\text{ID}}$. In the open-set recognition (OSR) problem, the objective is to maintain high classification accuracy on ID samples while improving the detection and handling of OOD samples. Our approach focuses on the semantic shift in the embedding space to detect OOD instances. Specifically, we estimate the probability \(P(y_j = \text{OOD} \mid z_j^{\text{dec}}, c_j)\), where \(z_j^{\text{dec}}\) is the embedding passed to the decoder and \(c_j \in \{1, 2, \dots, K_c\}\) is the assigned cluster. The number of optimal clusters \(K_c\) is not known a priori but is inferred through the self-supervised structure of \dynaSubVAE. 

For simplicity, throughout this paper, capital letters denote batched data, while lowercase letters refer to individual samples.

\paragraph{Model Workflow: }The overall structure of \dynaSubVAE\ is illustrated in Figure~\ref{fig:method_overview}. The input $X$ with batch size $B$, is first passed through the encoder $Enc(X)$ to produce a low-dimensional embedding \( H \in \mathbb{R}^{B \times D1} \), where \( D1 \) is the embedding dimension. The clustering embedding $Z_c$ is then sampled from a Gaussian posterior with parameters estimated by $H$. 
% The embedding \( H \) is then mapped to clustering embeddings \( Z_c \) using the reparameterization trick, allowing it to be sampled from a Gaussian distribution. 
Based on the parameters of the GMM-based clustering, namely the component means \( \boldsymbol{\eta_{\mu}} \in \mathbb{R}^{K_s \times D_2} \), standard deviations \( \boldsymbol{\eta_{log\sigma}} \in \mathbb{R}^{K_s \times D_2} \), and prior probabilities \( \boldsymbol{\pi} \in \mathbb{R}^{K_s} \), the embeddings \( Z_c \) are assigned to their corresponding subgroups. The clusters $C$ later help define the confidence in the subgroups and the final OOD flags~\ref{subsubsection:regret_func}. We assume an initial number of clusters \( K_s \), where \( K_s \ll N \), with \( N \) being the full input size. The number of clusters $K$ will be adapted during training. 

The intermediate embedding \( Z \), drawn from a Gaussian posterior parameterized by $H$, is passed through the adaptive modulation function \( f^{am}(Z, C) \), resulting in the final embedding \( Z_{\text{dec}} \), as shown in Equation~\ref{equ:Fam}. Here, \( f^{\text{inc}}_{C} \) is an incremental learning layer that trains and updates subgroup-specific weights and biases based on the input data. We perform feature modulation of the embedding through adaptive scaling based on the subgroup \( g_{W \mid C} \). To enable incremental and conditional updates of embeddings tailored to each subgroup, an additive residual interaction function \( f^{\text{inc}}_{C}(Z) \) is introduced as a subgroup-specific residual.
The decoder function \( \text{Dec}(Z_{\text{dec}}) \) then takes \( Z_{\text{dec}} \) and reconstructs the original image. To enhance the robustness of the VAE structure, we randomly crop partial regions of the original image during training.

\begin{align}
    Z_{dec} &= f^{am}(Z,C) = g_{W|C} * Z + f^{inc}_{C}(Z) \\
    g_{W|C} &= \sqrt{\text{softplus}(W_{C_k})} \;\;\;\;
\text{where} \quad \text{softplus}(x) = \log(1 + e^x)
    \label{equ:Fam}
\end{align}

\subsection{Dynamic Subgrouping}
\label{subsection:dynamic_subgrouping}

Through subgrouping, participants are divided into groups that exhibit more homogeneous behavior in the embedding space compared to those from different groups. This is achieved using a dynamic, centroid-based clustering approach inspired by Gaussian Mixture Models.

In GMM-based clustering tasks, forcing the approximate posterior $q(Z|X)$ to closely match a simple prior can disrupt the latent cluster structure. The KL divergence term may cause all data points to collapse toward the origin, leading to similar latent representations and a loss of meaningful cluster separation, known as the anticlustering effect \cite{dilokthanakul2016deep}. To mitigate this, we employ the intermediate low-dimensional embedding $H$ instead of $X$.
%before it is mapped to the mean $\mu$ and log-variance $\log(\sigma)$ used in the reparameterization trick. 
This $H$ serves as the input to the dynamic subgrouping module. Within this module, we compute $\mu_c$ and $\log(\sigma_c)$, and apply the reparameterization trick to obtain $Z_c = \mu_c + \epsilon \cdot \exp(\log(\sigma_c))$, where $\epsilon \sim \mathcal{N}(0, I)$. This yields a latent representation $Z_c \in \mathbb{R}^{B \times D_2}$ in a lower-dimensional space, which facilitates learning Gaussian parameters $\boldsymbol{\eta}_{\mu}$ and $\boldsymbol{\eta}_{\log \sigma}$ for each subgroup. Importantly, the KL divergence loss is not applied to $Z_c$ in order to avoid the anticlustering effect and preserve the integrity of the latent cluster structure.

The dynamic subgrouping model jointly optimizes the parameters that generate $Z_c$ and the clustering parameters $\theta$, in order to determine the most likely cluster assignment $\hat{C}$, as defined in Equation~\ref{equ:chat}, which represents the data embedding. 
\begin{equation}
    \hat{c} = \arg\max_k \log p(Z_c \mid c_k; \theta), \quad
    \text{where }\; \theta = \left\{ \boldsymbol{\eta}_{\mu}, \boldsymbol{\eta}_{\log \sigma}, \boldsymbol{\pi} \right\}
\label{equ:chat}
\end{equation}

% the number of clusters, the clustering algorithm and the metric or the similarity structure
\begin{algorithm}[hbt!]
\caption{Adaptive clustering algorithm.}
\label{alg:clustering_update}
\begin{algorithmic}
\STATE \textbf{Input:} $Z_c$ (latent representations), $K_s$ (starting cluster count)

\STATE \textbf{Initialize:} $K \gets K_s$, cluster parameters $\{\boldsymbol{\eta}_\mu^k, \boldsymbol{\eta}_{\log \sigma}^k, \boldsymbol{\pi}^k\}_{k=1}^{K}$, silhouette score $\text{Score}_{\text{Sil}} \gets \text{None}$
\STATE \textbf{Function: }{$f_{\text{update}}(Z_c^{(k)})$: 
 } $\boldsymbol{\eta}_{\log \sigma}^K = 0.095$; $\boldsymbol{\pi}^K = 0.001$; $\boldsymbol{\eta}_\mu^K = $ Sample far from exciting \{$\boldsymbol{\eta}_\mu^k\}_{k=1}^{K-1}$
\REPEAT
    \IF{iteration \% 100 == 0 \hfill \#Impose a constraint on cluster addition frequency}
        \IF{Epoch $\ne 0$}
            \IF{$\text{Score}_{\text{Sil}} < 0.5$ \hfill \#Decide whether embedding are not well separated by clusters } 
                \STATE $K \gets K + 1$ and Add new cluster via $f_{\text{update}}(Z_c^{(K)})$ 
            \ENDIF
        \ELSIF{$\frac{1}{K} \sum_{k=1}^K variance(Z_c^{(k)}) > 1.5$}
            \STATE $K \gets K + 1$ and Add new cluster via $f_{\text{update}}(Z_c^{(K)})$
        \ENDIF
    \ENDIF
    \STATE \textbf{Each Epoch: }Compute number of samples per cluster: $N_c^{(k)}$
    \IF{$\max_k N_c^{(k)} > 0.4 \sum_{j \ne k} N_c^{(j)}$ \hfill \#Split the dominant cluster}
        \STATE \textbf{Train} KMeans($Z_c^{(k)}$, $n_{\text{clusters}}=2$) to obtain cluster centers $M_1$ and $M_2$
        \STATE \textbf{Update} cluster k parameters $\boldsymbol{\pi}^k = \boldsymbol{\pi}^k/2$; $\boldsymbol{\eta}_\mu^k = M_1$
        \IF{There exists a cluster $C_u$ such that the embeddings $Z_c$ have moved away from it after several epochs, and it was previously assigned but is now unused}
            \STATE $\boldsymbol{\eta}_\mu^u = M_2$; $\boldsymbol{\pi}^u = \boldsymbol{\pi}^k$; $\boldsymbol{\eta}_{\log \sigma}^u = \boldsymbol{\eta}_{\log \sigma}^k$
        \ELSE
            \STATE $K \gets K + 1$
            \STATE $\boldsymbol{\eta}_\mu^K = M_2$; $\boldsymbol{\pi}^K = \boldsymbol{\pi}^k$; $\boldsymbol{\eta}_{\log \sigma}^u = 0.095$
        \ENDIF
        
    \ENDIF
\UNTIL{VAE loss converges}
\STATE \textbf{Output:} Optimal number of clusters $K_c \gets K$
\end{algorithmic}
\end{algorithm}

The algorithm for dynamically splitting or adding new cluster centers is described in Algorithm~\ref{alg:clustering_update}. Detailed explanations of each function, along with the formula used for merging cluster centers, are provided in the supplementary material.

In this algorithm, we use the silhouette score \cite{rousseeuw1987silhouettes} which is calculated as the average of individual point scores, defined as $\frac{\max(a_i, b_i)}{(b_i - a_i)}$. For a given point $Z_{c_i}$, $a_i$ denotes the average intra-cluster distance (to other points in the same cluster), while $b_i$ represents the average distance to the nearest neighboring cluster (inter-cluster distance). This score guides whether new clusters should be introduced to improve subgroup representation.

\subsubsection{Dynamic Subgrouping Losses}

The dynamic subgrouping loss comprises several components: the estimated ELBO loss ($\mathcal{L}_{NLL} + \mathcal{L}_{\text{KL}}$), a splitting loss ($\mathcal{L}{\text{split}}$) to control the variance within clusters, an entropy loss ($\mathcal{L}_{\text{entropy}}$) to ensure cluster assignments carry informative content about the embeddings, a usage entropy loss ($\mathcal{L}_{\text{usage}}$) and KL balancing loss ($\mathcal{L}_{KL\_balance}$) to promote balanced usage across clusters, and an augmentation loss ($\mathcal{L}_{\text{aug}}$) to encourage consistent clustering of original and augmented data. The final dynamic subgrouping loss is computed as a weighted sum of these terms:

\begin{equation}
\mathcal{L}_{\text{total}} = 
\lambda_{\text{elbo}} \left( \mathcal{L}_{\text{NLL}} + \mathcal{L}_{\text{KL}} \right) 
+ \lambda_{\text{split}} \mathcal{L}_{\text{split}} 
+ \lambda_{\text{ent}} \mathcal{L}_{\text{entropy}} 
+ \lambda_{\text{usage}} \mathcal{L}_{\text{usage}} 
+ \lambda_{\text{KLb}} \mathcal{L}_{\text{KL\_balance}} 
+ \lambda_{\text{aug}} \mathcal{L}_{\text{aug}}
\end{equation}

In the loss computations, we incorporate the log-posterior derived in Eq. \ref{equ:qc}. To convert these log posterior values into soft assignments (posterior probabilities over clusters), we apply the softmax function, with a temperature parameter $\tau$ for annealing.  

\begin{equation}
\log q_c^{(k)} = \log p(Z_c \mid c_k; \theta) + \log \pi_k
\label{equ:qc}
\end{equation}

% \begin{equation}
% \log q_y = \log p(Z_c \mid C; \theta) + \log \boldsymbol{\pi}, \quad \text{then} \quad q_y = \text{softmax}\left( \frac{\log q_y}{\tau} \right)
% \end{equation}

\paragraph{KL Divergence Loss and Negative Log Likelihood (NLL):} 
Let $\log p_c^{(k)}$ denote the log softmax of the cluster prior weights $\pi$. The KL divergence between the assignment distribution $q_c$ and the prior $p_c$ is given by Eq. \ref{equ:kldiv}. This loss encourages the learned cluster assignment distribution to stay close to the prior distribution defined by the mixture weights $\pi$. The summation of $\mathcal{L}_{NLL}$ loss and $\mathcal{L}_{\text{KL}}$ provides a lower bound for ELBO loss \cite{jordan1999introduction}. 

\begin{equation}
\mathcal{L}_{\text{KL}} = \mathbb{E}_{q_c} \left[ \sum_{k=1}^{K} q_c^{(k)} \left( \log q_c^{(k)} - \log p_c^{(k)} \right) \right] \quad \text{and} \quad \mathcal{L}_{NLL} = - \log (\sum_{k=1}^{K} \pi_k \dot p(Z_c \mid c_k; \theta))
\label{equ:kldiv}
\end{equation}
 
\paragraph{Splitting Loss:} 
To encourage better separation of clusters, we introduce the splitting loss, Eq. \ref{equ:splitting}, that penalizes clusters with high internal variance. The idea is to prevent overly diffuse clusters by comparing each cluster's variance to a global threshold. Where $\tau_2$ is a scaling factor, $\text{Var}_i$ is the variance of cluster $i$, and $\text{Var}_{\text{total}}$ is the variance of the entire dataset.

%\paragraph{Separation penaly:} This term penalizes cluster centers that are too close to each other. It computes pairwise distances between cluster means and adds a penalty to the loss if any distance falls below a threshold 2. 

\begin{equation}
\mathcal{L}_{\text{split}} = \sum_{i=1}^{K} \max\left(0, \text{Var}_i - \tau_2 \cdot \text{Var}_{\text{total}} \right)
\label{equ:splitting}
\end{equation}

\paragraph{Loss entropy: } 
This loss encourages low-entropy (i.e., confident) assignments by penalizing uncertain (high-entropy) distributions over clusters for individual samples.
\begin{equation}
\mathcal{L}_{\text{entropy}} = - \mathbb{E}_{q_c} q_c^{(k)} \log q_c^{(k)}
\end{equation}

\paragraph{Cluster Usage Entropy Loss: }  To encourage balanced usage, this loss measures the entropy of the overall distribution of cluster usage and is maximized when all clusters are used equally. 

\begin{equation}
\mathcal{L}_{\text{usage}} = - \sum_{k=1}^K u_k \log (u_k + \epsilon), \quad \textit{where} \quad u_k = \frac{1}{N} \sum_{i=1}^N q_{c_i}^{(k)}
\end{equation}

\paragraph{KL Balancing Loss: } 
This loss regularizes the model to distribute cluster assignments uniformly across all clusters. It minimizes the KL divergence between the average predicted cluster distribution $\mathbb{E}_{q_c} q_c^{(k)}$ and a uniform prior. By doing so, it encourages balanced usage of clusters and prevents mode collapse. 

\paragraph{Augmentation loss: } 
To promote domain-invariant clustering, we introduce an augmentation loss that aligns the cluster assignments of original and augmented versions of the same input. Specifically, we minimize the KL divergence between the soft cluster assignment of the augmented input and that of its original counterpart:

\begin{equation}
\mathcal{L}_{\text{aug}} = \text{KL}\left( q_c^{\text{aug}} \,\|\, q_c^{\text{orig}} \right) = \sum_{k=1}^K q_c^{\text{aug}, (k)} \left( \log q_c^{\text{aug}, (k)} - \log q_c^{\text{orig}, (k)} \right)
\end{equation}

\subsubsection{Regret Function}
\label{subsubsection:regret_func}
The subgroups capture distinct regions of the embedding space learned by the model. As new data arrive, some samples may not align well with existing subgroups, indicating potential misrepresentation by the VAE or the classifier. To address this, we introduce a regret function, defined in Equation~\ref{equ:regret_equation}, which quantifies confidence in subgroup assignments by measuring the change in prediction loss resulting from alternative subgroup assignments. Here, the loss refers to the cross-entropy between the pseudo-label and the predicted logits. The pseudo-label is derived from current label assignments using a classifier not trained on OOD data, allowing the model to assess uncertainty in subgroup assignment and the performance of alternative clusters. Subgrouping is performed in the embedding space, as input that seem distant under metrics like Manhattan distance may still fall within the model's generalization capacity. This highlights the importance of using latent representations over raw inputs for effective OOD detection.
\begin{equation}
R_c = \max_{c' \neq c} \left[ \mathcal{L}(p(y|c)) - \mathcal{L}(p(y|c')) \right] + \text{margin}
\label{equ:regret_equation}
\end{equation}

%This loss encourages the model to assign similar cluster probabilities to both original and augmented versions of the same image, enhancing the consistency and robustness of the clustering across input transformations.

% For upper bound calculation: 
% \begin{equation}
%     L(\theta, \hat{\theta}) = \int (\theta-\hat{\theta}(x))^2 \,dx 
% \end{equation}
% \begin{equation}
%     \mathbb{R}^n \equiv \mathbb{R}^n(P) = \inf_{\hat{\theta}} \sup_{P \in \mathcal{P}} \mathbb{E}_P[d(\hat{\theta}, \theta(P))]
% \end{equation}

% \begin{equation}
%     \mathbb{R}^n \leq \sup_{P \in \mathcal{P}} \mathbb{E}_P[d(\hat{\theta}, \theta(P))] = U_n
% \end{equation}
% \begin{equation}
%      L_u = t \inf_{\hat{\theta}} \sup_{P \in \mathcal{P}} P(d(\hat{\theta}, \theta(P))<t) \leq \mathbb{R}^n 
% \end{equation}

% \begin{equation}
%     \theta* = \mathbb{E}[Z|X1] = \mathbb{E}[\mathbb{E}[Z|X1,Y]|X1]
% \end{equation}

% \begin{equation}
%     \theta* =\mathbb{E}[\mathbb{E}[Z|Y]|X] = \sum_{y} P(Y=y|X)\mathbb{E}[Z|Y=y]
% \end{equation}

% \begin{equation}
% \text{minimize } l \text{given }
%     \label{equ:opt}
% \end{equation}

\subsection{Representation Learning Losses}
The VAE loss for representation learning includes four components: reconstruction loss ($L_{\text{recon}}$) for accurate image reconstruction, KL divergence loss ($L_{\text{kl}}$) to enforce a normal latent distribution, contrastive loss ($L_{\text{contrast}}$) to align embeddings of similar images, and orthogonality loss ($L_{\text{ortho}}$) to ensure subgroup-specific weights contribute to learning distinct and informative features. The final VAE loss is expressed as a weighted sum of these components:

\[
\mathcal{L}_{\text{VAE}} = \beta_{\text{recon}} \mathcal{L}_{\text{recon}} + \beta_{\text{kl}} \mathcal{L}_{\text{kl}} + \beta_{\text{contrast}} \mathcal{L}_{\text{contrast}} + \beta_{\text{ortho}} \mathcal{L}_{\text{ortho}},
\]

% where $\beta_{\text{recon}}$, $\beta_{\text{kl}}$, $\beta_{\text{contrast}}$, and $\beta_{\text{ortho}}$ are the respective weights for each loss term.

\label{subsection:VAElosses}
\paragraph{Reconstruction loss: } The mean squared error (MSE) between the reconstructed image and the original (non-cropped) image is used as the reconstruction loss. In addition, a Kullback–Leibler (KL) divergence loss is applied to encourage the latent embeddings $Z$, derived via the reparameterization trick from $\mu$ and $\sigma$, to approximate a standard normal distribution.

\paragraph{Contrastive Loss:} We employ a contrastive loss strategy that brings augmented data samples, when the augmentation preserves domain identity, closer together in the embedding space. This approach aligns with domain-aware augmentation techniques, as demonstrated by Dubois et al.~\cite{dubois2021optimal}, which highlight the need for partial domain knowledge to enable accurate domain shift detection. To enforce separation across domains, we divide labels into 3 distinct groups. For each data point, we identify samples within the batch that belong to a different group and treat them as negative examples to push away in the embedding space. If no such negative exists in the batch, we use a standard noise embedding as a surrogate negative sample. Conversely, augmented versions of a data point, or samples from the same group, are treated as positive examples for triplet loss computation \cite{schroff2015facenet}.

\paragraph{Orthogonality Loss:}
The orthogonality loss ensures that the latent representations with clustering ($Z^{dec}$) and without clustering ($Z$) remain distinct, as defined in Eq.~\ref{equ:orth_loss}. This loss encourages $Z$ and $Z^{dec}$ to be orthogonal by discouraging cosine similarity \cite{singhal2001modern} between two embeddings, thereby reducing redundancy in the latent space. It ensures that the effect of cluster assignment is reflected in the latent representation and subsequently accounted for during reconstruction.

\begin{equation}
    \mathcal{L}_{\text{ortho}} = \text{mean} \left( \left| \text{cosine\_similarity}(Z, Z^{dec}) \right| \right)
    \label{equ:orth_loss}
\end{equation}

\section{Experiments}
\paragraph{Datasets:}  
We evaluate our approach on 3 simulated and 3 real-world datasets. For the simulated data, we use six labeled classes and intentionally omit one label during training to simulate out-of-distribution (OOD) conditions. We consider three synthetic datasets: 1. A well-separated Gaussian blob dataset \cite{pedregosa2011scikit}, structured to ensure 100\% separability between clusters. This setup allows us to evaluate our model's ideal-case OOD detection performance. 2. A Moon dataset with nonlinear separability, introducing complexity by integrating overlapping regions. This presents a more challenging scenario for clustering and OOD detection. 3. A Circles dataset with concentric circular patterns that are not linearly or Gaussian-separable. This serves as a difficult case, allowing us to test whether the model can embed the data into a space where clusters become Gaussian-like and OOD detection remains reliable. In addition to synthetic data, we evaluate \dynaSubVAE\ on several real-world datasets: \textbf{MNIST} \cite{lecun1998gradient}: A widely used handwritten digit dataset with 10 classes. \textbf{CIFAR-10} \cite{krizhevsky2009learning}: A 10-class dataset consisting of natural images across various object categories. \textbf{MedMNIST}~\cite{medmnistv1, medmnistv2} is a large-scale collection of lightweight medical image datasets covering a variety of tasks and imaging modalities. In our experiments, we specifically use the \textbf{PathMNIST} subset, which consists of colorectal cancer histology images categorized into 9 classes representing both tumor-related (cancerous) and non-tumor (benign) tissue types. 

\paragraph{Training Schema:}  
For both the subgroup module and VAE training, the Adam optimizer was used with a learning rate of $5 \times 10^{-4}$ and a weight decay of $1/500$. Cosine annealing learning rate scheduler~\cite{loshchilov2016sgdr}, with a maximum number of iterations $T_{\text{max}} = 200$ was employed. The integrated subgroup model begins updating two epochs after the VAE model starts training. Detailed explanations of the loss weighting coefficients and hidden dimension sizes are provided in the supplementary material.
For the simulated datasets, a single-layer linear logistic regression classifier was sufficient to predict class labels from $Z^{dec}$ with high accuracy. In the case of real-world datasets, a simple two-layer neural network with a ReLU activation function between the layers achieved reliable performance. Classifier training was performed using cross-entropy loss and the SGD optimizer with a learning rate of 0.01 and a batch size of 32. \dynaSubVAE\ is resource-friendly, with training requiring only a GPU with 120GB of memory. 
% All model hyperparameters, including loss weights, are provided in the supplementary material. 

\paragraph{Evaluation Metrics:} 
We assessed performance using several metrics. \textbf{ID accuracy} measures classification performance on known classes by aligning clusters to ground-truth labels. \textbf{OOD accuracy} quantifies the proportion of out-of-distribution samples correctly identified as OOD. \textbf{Normalized Mutual Information (NMI)} and \textbf{Adjusted Rand Index (ARI)} \cite{hubert1985comparing} evaluate clustering quality by comparing predicted clusters with true labels, where NMI captures mutual dependence and ARI adjusts for chance alignment. For OOD detection benchmarking, we reported \textbf{AUROC} (area under the ROC curve), which measures the model’s ability to distinguish ID from OOD samples across thresholds, and \textbf{FPR} (false positive rate at 95\% true positive rate), which indicates the error rate when high sensitivity is required.

% \subsection{Convergence Analysis}
% The classes that has regret value considering Equ. \ref{equ:regret_class} is more than 0.7 and their classification prediction confidence is less than 50\%. 
% The percentage of the time that number 9 is detected to be out of distribution is 99.23\% compared to other classes. 

% The convergence analysis typically involves showing that the centroids $C_{t-1}^p$ converge to the true centroids as t approaches infinity. This is often demonstrated by showing that the expected value of the squared distance between the online centroids and the optimal centroids decreases over time.

% As this the self-supervised learning approach the true subgroups where the data is belonged to is not known. Therefore, we experimentally shows regret performance for both simulated and real-world datasets.
\subsection{Results}
We evaluated OOD performance in three settings: near-OOD (e.g., MNIST vs. FashionMNIST \cite{xiao2017fashion}), far-OOD (e.g., MNIST vs. CIFAR-10 or SVHN \cite{netzer2011reading}), and class-OOD, where the OOD samples are unseen classes from the same dataset; making them especially hard to detect due to high distributional similarity.

Table \ref{tab:dataset_class_ood} illustrates the effectiveness of the subgroup-based regret function in detecting class-wise OOD samples. For this analysis, one class was randomly dropped during training, and the OOD accuracy reflects the model’s ability to detect the omitted class. In our framework, maintaining high accuracy on in-distribution (ID) samples is also essential. As shown in the table, the ID accuracy remained high across all datasets.

\begin{table}[!ht]
    \centering
    \begin{tabular}{l|cccc}
        Dataset & ID-Accuracy $\uparrow$ & Class OOD Accuracy $\uparrow$ & NMI $\uparrow$ & ARI $\uparrow$\\
        \toprule
        %Simulated data Circles (3 - 5) &  69\% & 0.75 & 0.55\\
        Simulated data Circles & 99\% & 91\% & 0.84 & 0.72\\ %(2)
        % currect regreted preds / len(all regreted preds): 0.84
        % Input distance comparison 41\% accuracy
        % 4 is 100% accurate
        Simulated data Blobs & 100\% & 98\% & 0.96 & 0.94\\ %(2)
        %Simulated data Blobs (4) & 100\% & 0.90 & 0.79\\ %like this one more and with flow
        Simulated data Moons & 100\% & 98\% & 0.74 & 0.60\\ % (1)
        % currect regreted preds / len(all regreted preds): 0.95
        % Input distance comparison 47\% accuracy
        Mnist & 96\% & 86\% &  0.52 &  0.34 \\ % (4)
        %currect regreted preds / len(all regreted preds):  0.7 
        %85\% if all labels
        MedMnist & 92\%  & 97\% & 0.71 & 0.49\\ % (2)
        % 67\% - 84-91 for (3)
        % Cifar10 (1-4 resnet18) & 73\% & 74\% &  & \\
        %0.26
    \end{tabular}
    \caption{Class-wise OOD evaluation on simulated data, MNIST, and MedMNIST datasets.}
    \label{tab:dataset_class_ood}
\end{table}

We comprehensively compared the near and far OOD performance of state-of-the-art (SOTA) models and \dynaSubVAE\ in Table~\ref{tab:sota_compare}. In this table, “Acc” refers to OOD accuracy. On far OOD detection tasks, our model outperformed all other methods. For near OOD detection, its performance ranked among the top 2–3 models, which is still competitive. Notably, all other compared methods are fully supervised, whereas \dynaSubVAE\ relies on a simple MLP classifier trained on learned embeddings. While its ID accuracy was slightly lower, it maintained strong OOD detection performance.

\begin{table*}
% \scriptsize
\footnotesize
\scriptsize
    \centering
    \resizebox{1.\textwidth}{!}{\begin{tabular}{lc|ccc|ccc|ccc}
        \multicolumn{2}{c}{Cifar10 (ID)} &\multicolumn{3}{c}{Cifar100 (Near OOD)}&\multicolumn{3}{c}{Mnist (Far OOD)}&\multicolumn{3}{c}{SVHN (Far OOD)}\\
        \toprule
        Models & \makecell{ID\\Acc $\uparrow$} & \makecell{OOD\\Acc $\uparrow$} & AUROC $\uparrow$ & FPR $\downarrow$ & \makecell{OOD\\Acc $\uparrow$} & AUROC $\uparrow$ & FPR $\downarrow$ & \makecell{OOD\\Acc $\uparrow$} & AUROC $\uparrow$ & FPR $\downarrow$ \\
        \midrule
        % logitnorm & 93.64\% & 73.7\% & 90.77 & 35.45 & 95.3\% & 98.92 & 4.59 & 89.3\% & 97.76 & 9.12\\
        KNN & 95.17\% &71.8\% &\textbf{89.43} & \textbf{39.84} & 78.8\% & 93.72 & 23.03 & 62.1\% & 92.27 & 21.38\\
        DICE & 94.91\% &63.3\% & 76.09 & 84.74 & 86.6\% & 95.16 & 26.71 & 69.7\% & 89.27 & 50.73\\
        Fdbd & 95.17\% &69.1\% & 87.46 & 57.47 & 75.2\% & 90.93 & 36.09 & 54.7\% & 88.77 & 36.90\\
        Scale & 95.17\% &\textbf{72.2\%} & 85.49 & 72.9 & 86.8\% & 95.51 & 18.06 & 70.0\% & 91.88 & 31.57\\
        MSP &  95.17\% & 68.0\% & 86.67 & 59.88 & 75.7\% & 93.20 & 21.13 & 67.3\% & 90.56 & 25.76\\
        \midrule
        % RODD &94.45\%&&&&&&&& 99.63 & 1.82\\
        % RODD (unsupervised) &94.45\%&&&&&&&& 99.63 & 1.82\\
        DynaSubVAE & 93.10\% & 64.1\% & 85.45 & 55.53 & \textbf{89.8\%} & \textbf{97.84} & \textbf{8.61} & \textbf{84.3\%} & \textbf{96.86} & \textbf{7.52}
        % class OOD is 56\% with 55\% accuracy 

    \end{tabular}}
    \caption{Near and far OOD detection performance compared to SOTA models on CIFAR-10 dataset.}
    \label{tab:sota_compare}
\end{table*}

For the analysis in Table~\ref{tab:cluster_comparison}, we replaced the trained \dynaSubVAE\ subgrouping model with external clustering methods, Kmean++ \cite{arthur2006k} and GMM \cite{dempster1977maximum}, and evaluated the resulting class OOD detection performance. Regret Precision measures the percentage of correctly predicted OOD samples out of all samples flagged as OOD. This comparison highlights the effectiveness of our dynamic subgrouping approach and underscores the importance of jointly training it alongside the VAE model. 

\begin{table*}
% \footnotesize
\small
    \centering
    % \resizebox{1.\textwidth}{!}
    {\begin{tabular}{l|cc|cc|cc}
        &\multicolumn{2}{c}{\dynaSubVAE}&\multicolumn{2}{c}{GMM}&\multicolumn{2}{c}{Kmean++}\\
        \toprule
        Dataset & OOD  & Regret  & OOD  & Regret  & OOD  & Regret \\
         &  Accuracy &  Precision &  Accuracy &  Precision &  Accuracy &  Precision\\
        \toprule
         Simulated Circles &  99\% & \textbf{98\%} & 80\% & 50\% & \textbf{100\%} & 54\%\\
         %90\% accuracy if train classifier on all data but not the VAE
         Simulated Moons & \textbf{100\%} & \textbf{95\%} & 55\% & 82\% & \textbf{100\%} & 85\%\\
         %99\% accuracy if train classifier on all data but not the VAE
         Simulated Blobs & \textbf{98\%} & \textbf{100\%} & 23\% & \textbf{100\%} & 96\% & 99\%\\
         %100\% accuracy if train classifier on all data but not the VAE
         Mnist & \textbf{86\%} & \textbf{72\%} & 46\% & 39\% & 44\% & 38\%\\
         % MedMnist & \textbf{92\%} & \textbf{71\%} & _ & _ & _ & _\\ 
    \end{tabular}}
    \caption{OOD performance comparison between \dynaSubVAE\ and external clustering techniques.}
    \label{tab:cluster_comparison}
\end{table*}

We performed continuous updating analysis by initiating a new subgroup for the OOD data once it reached at least 32 instances. We trained only the weights associated with the new subgroup and retrained the classifier based on the updated embeddings. This approach led to improvements in accuracy of 11\%, 17\%, and 15\% on the simulated Circles, Moons, and Blobs datasets, respectively.
% \begin{table*}
%     \centering
%     \begin{tabular}{c|c}
%     Methods & accuracy improvement after VAE fine tuning\\
%         \toprule
%          Simulated data Circles &  11\%\\
%          Simulated data Moons &  17\%\\
%          Simulated data Blobs &  15\%\\
%     \end{tabular}
%     \caption{Impact of clustering and OOD detection}
%     \label{tab:dynamic_update}
% \end{table*}

\subsection{Ablation Study}

Table \ref{tab:ablation_study} illustrates the key components of \dynaSubVAE\ and their impact on far, near, and class OOD detection. Without the orthogonal loss ($\mathcal{L}_{\text{ortho}}$), the clustering model failed to learn disentangled representations, leading to large gradient magnitudes and poor performance in class OOD detection. In the augmentation loss ($\mathcal{L}_{\text{aug}}$), we replaced the soft augmentation loss, based on KL divergence between the clustering posteriors of the original and augmented images, with a hard consistency loss that enforces exact cluster assignment matching. The results highlight the importance of each loss component for effective separation between OOD and ID samples.
\begin{table*}[h]
\small
    \centering
    \begin{tabular}{l|c|ccc|ccc}
         & Class OOD & \multicolumn{3}{c}{Fashion Mnist (Near OOD)}&\multicolumn{3}{c}{SVHN (Far OOD)}\\
          & Acc $\uparrow$ & Acc $\uparrow$ & AUROC $\uparrow$ & FPR $\downarrow$ & Acc $\uparrow$ & AUROC $\uparrow$ & FPR $\downarrow$\\
        \toprule
        wo $\mathcal{L}_{ortho}$ & 22\% &76.6\% & 93.66 & 20.95 &69.7\%& 89.09& 38.41\\
        wo $\mathcal{L}_{cont}$ & 32\% &76.5\%& 93.28& 22.85&70.1\%&88.44 & 45.12\\
        wo $\mathcal{L}_{KL\_{balance}}$ & 42\% & 77.4\% & 93.67 & 21.29 & 71.1\% & 88.97 & 40.96\\
        $\mathcal{L}_{aug}$ Soft$\rightarrow$Hard & 72\% & 77.0\% & 93.51 & 21.88& 68.6\% & 86.35 & 56.50\\
        \dynaSubVAE & \textbf{86\%} &  \textbf{79.6\%} & \textbf{94.54} & \textbf{18.00} & \textbf{92.5\%} & \textbf{97.08} & \textbf{9.23}\\        
    \end{tabular}
    \caption{Ablation study evaluating performance across different OOD types. ‘wo’ indicates ‘without’.}
    \label{tab:ablation_study}
\end{table*}

\section{Conclusion}

We proposed \dynaSubVAE, a non-parametric, dynamic subgrouping mechanism integrated into the VAE framework to identify clusters in the embedding space and address a regret-based objective for OOD detection. The novel architecture allows partial updates to the embedding layer and supports the dynamic expansion of cluster numbers as new OOD data arrives, enabling better representation of previously emerging data. In our experiments, \dynaSubVAE\ demonstrated competitive performance in both near and far OOD detection, even compared to fully supervised methods. On the CIFAR-10 dataset, our method reduced FRP@95 by 29\% and 9.45\%, and improved OOD detection AUROC by 4.59\% and 2.33\% on SVHN and Mnist respectively, in the context of far-OOD detection compared SOTA models. Furthermore, our robust clustering strategy improves the detection of class-level OOD samples, a particularly challenging task in the current literature.

\textbf{Limitations and Future Work:} 
Our approach involves multiple loss terms to ensure effective integration between clustering and reconstruction, which introduces additional complexity in hyperparameter tuning. 
% While all loss components were necessary in our experiments, future work could explore simplification by reducing the number of loss terms. 
Although \dynaSubVAE\ operates in a self-supervised setting and does not rely on labeled data, a more thorough evaluation under label noise conditions in OOD detection remains an important next step. Additionally, applying the integrated clustering mechanism to other unsupervised architectures such as GANs or diffusion models presents a promising direction for future research. 

%Already trained embeddings without fine-tuning as in this model the training of both is integrated and it takes the embedding to the dimension that is separable by GMM clusters. If the clustering models need to be trained separately, ensemble of different clustering methods would be very useful. 

% Future work: Improving the memory usage and instead computing distance for each new assignment find a way to derive upper bound and lower bound (hard as embedding change and clusters are applied to the embedding)

\begin{ack}
Resources used in preparing this research were provided, in part, by the Province of Ontario, the Government of Canada through CIFAR, and companies sponsoring the Vector Institute \href{the Vector Institute}{https://vectorinstitute.ai/partnerships/current-partners/}. T.B. acknowledges support from the Natural Sciences and Engineering Research Council of Canada (NSERC) through a PGS-D award.
\end{ack}

\newpage
\bibliographystyle{plainnat}
% \bibliography{ref}

\newpage
\appendix

\section{Methodology Further Details}
\subsection{Adaptive Clustering Strategy}

This part explains Algorithm 1 of main paper in further detail.
To ensure flexibility in representing the latent structure of the data, we introduce an adaptive clustering strategy that dynamically adjusts the number of clusters during training. This mechanism monitors clustering quality and decides whether to increase or decrease the number of clusters ($K$), based on either a silhouette score or intra-cluster variance. The adjustment allows the model to avoid both underfitting (too few clusters) and overfitting (too many), supporting both improved generalization and robust latent representation.

\paragraph{Silhouette-Based and Variance-Based Adjustment.} At each iteration, the model evaluates the current clustering configuration. If a silhouette score is available and is below a given threshold ($\text{sil\_score} < 0.5$), this suggests poor cluster separation, prompting an increase in the number of clusters.
Alternatively, if the silhouette score is not available (during the first epoch or when the number of detected subgroups in the previous epoch is fewer than the current number of clusters), the model calculates the mean intra-cluster variance:
\[
\text{Var}_{\text{mean}} = \frac{1}{K} \sum_{k=1}^K \text{Var}(Z_c^{(k)})
\]
If $\text{Var}_{\text{mean}}$ exceeds a predefined threshold $\tau$, the number of clusters is incremented by one. After evaluating several values $\tau \in \{1.2, 1.5, 1.7, 2\}$, we found that $\tau = 1.5$ yielded the best results. Whenever the number of clusters changes, the model invokes a parameter update routine to either initialize new clusters or merge existing ones.

\paragraph{Cluster Parameter Update Function:} The parameter update logic is handled by the function $f_{\text{update}}(Z_c^{(k)})$, which either adds a new cluster or merges existing ones. When a new cluster is added, its parameters are initialized as follows: 

\begin{itemize}
    \item $\boldsymbol{\eta}_{\log \sigma}^{K}$ is the log-variance of the new cluster, initialized to $\log(1.1^2) \approx 0.095$.
    \item $\boldsymbol{\pi}^{K}$ is the raw mixture weight for the new cluster, set to a small value to minimize its initial influence.
    \item $\boldsymbol{\eta}_\mu^K$ is the mean of the new cluster, sampled far from existing cluster centers $\left(\left\{\boldsymbol{\eta}_\mu^k\right\}_{k=1}^{K-1}\right)$ to ensure representational diversity.
\end{itemize}

\subsection{Merging clusters}
The decision to merge clusters is based on the symmetric Kullback–Leibler (KL) divergence \cite{hershey2007approximating} between their corresponding Gaussian components. This divergence metric quantifies how different two Gaussian distributions are, taking into account both their means and variances. Unlike simple distance metrics such as Euclidean distance, the symmetric KL divergence is distribution-aware, making it particularly suitable for probabilistic clustering scenarios. 
This metric captures: 1. Mean mismatch: Larger differences in means yield higher divergence, especially when the variances are small. 2. Variance mismatch: High divergence also occurs when one distribution is sharply peaked (low variance) while the other is broad (high variance), even if their means are close.
Due to its sensitivity to both location and uncertainty, symmetric KL divergence is well-suited for guiding GMM-based cluster merging, where clusters may have overlapping shapes.

We first identify all active clusters from the current epoch. For each unique pair of active clusters $(i, j)$, where $i < j$, we compute the symmetric KL divergence between their corresponding Gaussian components:
\[
\text{KL}_{\text{sym}}(\mathcal{N}_i \| \mathcal{N}_j) = \frac{1}{2} \left[ \text{KL}(\mathcal{N}_i \| \mathcal{N}_j) + \text{KL}(\mathcal{N}_j \| \mathcal{N}_i) \right],
\]
where $\mathcal{N}_i = \mathcal{N}(\boldsymbol{\eta}_{\mu_i}, \boldsymbol{\eta}_{\sigma_i})$ and $\mathcal{N}_j = \mathcal{N}(\boldsymbol{\eta}_{\mu_j}, \boldsymbol{\eta}_{\sigma_j})$. These values are stored for further comparison.

After incorporating the GMM parameters, the expression becomes:
\begin{align*}
\text{KL}_{\text{sym}}(\mathcal{N}_i \| \mathcal{N}_j) = \frac{1}{2} \sum_{d=1}^D \Bigg[ & 
\frac{\boldsymbol{\eta}_{\sigma_i}^{2(d)} + (\boldsymbol{\eta}_{\mu_i}^{(d)} - \boldsymbol{\eta}_{\mu_j}^{(d)})^2}{\boldsymbol{\eta}_{\sigma_j}^{2(d)}}
+ \frac{\boldsymbol{\eta}_{\sigma_j}^{2(d)} + (\boldsymbol{\eta}_{\mu_i}^{(d)} - \boldsymbol{\eta}_{\mu_j}^{(d)})^2}{\boldsymbol{\eta}_{\sigma_i}^{2(d)}} - 2 \\
& + \log\left(\frac{\boldsymbol{\eta}_{\sigma_j}^{2(d)}}{\boldsymbol{\eta}_{\sigma_i}^{2(d)}}\right)
+ \log\left(\frac{\boldsymbol{\eta}_{\sigma_i}^{2(d)}}{\boldsymbol{\eta}_{\sigma_j}^{2(d)}}\right)
\Bigg]
\end{align*}

To avoid merging well-separated clusters, we determine a dynamic threshold based on the distribution of computed distances. In practice, two Gaussians with a symmetric KL divergence below a certain threshold (e.g., 0.5 or \( 0.1 \times D \)) may be considered sufficiently similar to merge. We then sort all cluster pairs by divergence score and select the first pair $(i, j)$ with $\text{KL}_{\text{sym}}^{(i,j)} < \text{KL}_{\text{threshold}}$ as a candidate for merging.

Once a pair of clusters $(i, j)$ is selected, we merge them by computing the new parameters as a weighted average of their means:
\[
\boldsymbol{\eta}_{\mu_{\text{new}}} = \frac{{\pi_i} \boldsymbol{\eta}_{\mu_i} + {\pi_j} \boldsymbol{\eta}_{\mu_j}}{{\pi_i} + {\pi_j}}, \quad {\pi_{\text{new}}} = {\pi_i} + {\pi_j},
\]
where $\pi_i$ and $\pi_j$ are the mixture weights of clusters $i$ and $j$. The parameters of cluster $i$ are updated with $\boldsymbol{\eta}_{\mu_{\text{new}}}$ and $\pi_{\text{new}}$, while cluster $j$ is deactivated by setting its weight and variance to near-zero values:
\[
\pi_j \leftarrow 10^{-6}, \quad \boldsymbol{\eta}_{\log \sigma_j} \leftarrow 10^{-6}.
\]

\subsection{Mathematical Reasoning behind Adaptive Modulation}

How would the model break down in case of separate impact of subgroup: 

\begin{align}
    p(Z_{\text{dec}} \mid X) &= \sum_{c} p(Z_{\text{dec}}, c \mid X) = \sum_{c} p(c \mid X)\, p(Z_{\text{dec}} \mid c, X) \\
    \text{If } p(c \mid X) &= 1 \text{ for the optimal cluster assignment } \hat{c}, \text{ then:} \\
    p(Z_{\text{dec}} \mid X) &= p(Z_{\text{dec}} \mid \hat{c}, X)
\end{align}

Therefore, the mixture of Gaussians reduces to a single Gaussian corresponding to the assigned cluster $\hat{c}$. How we changed it:

We define the adaptive modulation as a transformation:
\[
E_{\text{mid}}^c = W_c Z + b_c + G_c H
\]

\begin{align}
    p(Z_{\text{dec}} \mid X) 
    &= \sum_{c} p\big(Z_{\text{dec}},c, Z, H, \mid X\big) \\
    &= \sum_{c} p(c, Z, H \mid X)\, p\big(Z_{\text{dec}} \mid c, Z, H, X\big) \\
    &= \sum_{c} p(c \mid Z, X, H)\, p(Z \mid X, H), p(H \mid X)\, p\big(Z_{\text{dec}} \mid c, Z, X\big)\\
    &= \sum_{c} p(c \mid H)\, p(Z \mid H)\, p(H \mid X)\, p\big(Z_{\text{dec}} \mid c, Z, X\big)
\end{align}

Moreover, the incorporated subgrouping losses are designed to prevent the model from collapsing all data into a single cluster, thereby avoiding the trivial solution of learning a single Gaussian embedding instead of a meaningful mixture.

\subsection{Regret Functions}

During testing, as new data arrives, the model evaluates whether the input belongs to the previously learned distribution. If the input is identified as OOD—based on the classes and subgroups defined by the dynamic subgrouping model, the system initiates the training of a personalized layer tailored to the new data. To guide this process, we use a regret function to quantify the performance gap between the current decision and a potentially better alternative.

In addition to the regret formulation defined in the main paper, we explored several alternative regret functions. These included variants based on reconstruction loss across different subgroups, as well as masking strategies using predicted class confidence entropy and probability thresholds, without conditioning on subgroup assignments. However, our proposed regret function—formulated as the difference in classification loss across subgroup classifiers, consistently outperformed the alternatives. It provided a more reliable measure of both confidence in subgroup representation and predictive accuracy, making it best suited for triggering adaptation in the OOD detection.

\section{Training and Structural Details}
\subsection{Model Details}
\paragraph{Encoder: }For the simulated data, a linear encoder is used. For the real-world datasets, a ResNet-18 encoder \cite{he2016deep, shafiq2022deep} is employed, with the first convolutional layer modified to accept 32×32 images. For the MNIST dataset, grayscale images are converted to 3-channel RGB-like format. The final classification layer of ResNet-18 is removed and replaced with a projection layer that maps to the $D_1$ hidden dimension. Two linear layers are used to map the embeddings to the mean $\mu$ and log-variance $\log(\sigma)$. The latent embedding $Z$ is then obtained using the reparameterization trick: $Z = \mu + \epsilon \cdot \exp\left(0.5 \cdot \log(\sigma)\right)$, where $\epsilon \sim \mathcal{N}(0, I)$.

\paragraph{Decoder: }For the simulated dataset, a simple linear decoder is employed to map the latent variable $Z$ directly back to the original input dimension. In contrast, for real-world image datasets, a more expressive decoder is used. Specifically, the latent embedding from the $D_1$-dimensional space is first projected through a fully connected layer into a tensor of shape $128 \times 4 \times 4$. This shape provides a compact spatial representation that facilitates progressive upsampling using transposed convolutional layers \cite{radford2015unsupervised}. The first layer increases the spatial dimensions from $4 \times 4$ to $8 \times 8$, followed by batch normalization and a ReLU activation. The second layer further upsamples to $16 \times 16$, again followed by normalization and activation. The final transposed convolutional layer expands the output to $32 \times 32$ with a number of channels matching the input image (e.g., 1 for MNIST, 3 for CIFAR). A sigmoid activation is applied at the end to ensure that the output pixel values lie within the $[0, 1]$ range.
This architecture enables the model to reconstruct high-resolution images from low-dimensional embeddings.

\paragraph{Subgrouping Model: } For the simulated dataset, a two-layer fully connected network with a ReLU activation in between is used to map the embedding from the $D_1$-dimensional space to $D_2$, passing through an intermediate dimension of $D_1/2$. For the real-world datasets, a single linear layer is used to map the representation $H$ from $D_1$ to $D_2$. The value of $D_2$ is fixed at 5 across all datasets to ensure that the multi-Gaussian representation lies in a lower-dimensional space, making it easier to learn and more stable to optimize. 

The mixture parameters, $\boldsymbol{\eta_{\mu}}$, $\boldsymbol{\eta_{\log \sigma}}$, and $\boldsymbol{\pi}$, are initialized to promote stable and diverse component representations at the start of training. For the first $K_s$ components, the means $\boldsymbol{\eta_{\mu}}$ are manually set to be evenly spaced between -1 and 1, encouraging diversity among initial clusters. The log-variances $\boldsymbol{\eta_{\log \sigma}}$ are initialized to $\log(1.1)$ to ensure strictly positive variances while maintaining numerical stability. The mixture weights $\boldsymbol{\pi}$ are parameterized using a softmax over values randomly drawn from a standard normal distribution, providing a valid initial probability distribution while preserving flexibility for optimization.

\subsection{Hyperparameters Tuning}
We performed a grid search on the MNIST dataset to derive suitable weights for the loss components. These weights were kept consistent across different datasets for simplicity. The selected weights for the subgrouping loss (see Equation \ref{equ:subgroup_losses}) are as follows: $\lambda_{\text{elbo}} = 1$, $\lambda_{\text{entropy}} = 3$, $\lambda_{\text{usage}} = 0.5$, $\lambda_{\text{KLb}} = 2$, $\lambda_{\text{split}} = 3$, and $\lambda_{\text{aug}} = 0.1$.

\begin{equation}
\mathcal{L}_{\text{total}} = 
\lambda_{\text{elbo}} \left( \mathcal{L}_{\text{NLL}} + \mathcal{L}_{\text{KL}} \right) 
+ \lambda_{\text{split}} \mathcal{L}_{\text{split}} 
+ \lambda_{\text{ent}} \mathcal{L}_{\text{entropy}} 
+ \lambda_{\text{usage}} \mathcal{L}_{\text{usage}} 
+ \lambda_{\text{KLb}} \mathcal{L}_{\text{KL\_balance}} 
+ \lambda_{\text{aug}} \mathcal{L}_{\text{aug}}
\label{equ:subgroup_losses}
\end{equation}

The loss weights used for the VAE total loss (see Equation~\ref{equ:vae_losses}) are: $\beta_{\text{recon}} = 1$, $\beta_{\text{kl}} = 1$, $\beta_{\text{contrast}} = 100$, and $\beta_{\text{ortho}} = 10$. For only simulated Circle data $\beta_{\text{kl}} = 0.2$.
\begin{equation}
\mathcal{L}_{\text{VAE}} = \beta_{\text{recon}} \mathcal{L}_{\text{recon}} + \beta_{\text{kl}} \mathcal{L}_{\text{kl}} + \beta_{\text{contrast}} \mathcal{L}_{\text{contrast}} + \beta_{\text{ortho}} \mathcal{L}_{\text{ortho}}
\label{equ:vae_losses}
\end{equation}

Subgroup-synchronized training begins at the second epoch for all datasets except CIFAR-10, where it starts at the fifth epoch. We evaluated multiple values of $D_1 \in \{32, 64, 80, 128, 256\}$ for each dataset. Based on these experiments, $D_1$ was set to 80 for the simulated dataset, 128 for MNIST and MedMNIST, and 256 for CIFAR-10.

\subsection{Training Loss Curves}
The validation set comprises 20\% of the training data. Early stopping is applied with a patience of 7 epochs, based on the validation loss. For evaluation, we use the best model checkpoint obtained before the validation loss begins to deteriorate. The validation loss consists solely of the reconstruction error, with reconstructed images conditioned on the predicted subgroup.

\begin{figure}[htbp]
    \centering
    \begin{subfigure}[b]{1\linewidth}
        \centering
        \includegraphics[width=\linewidth]{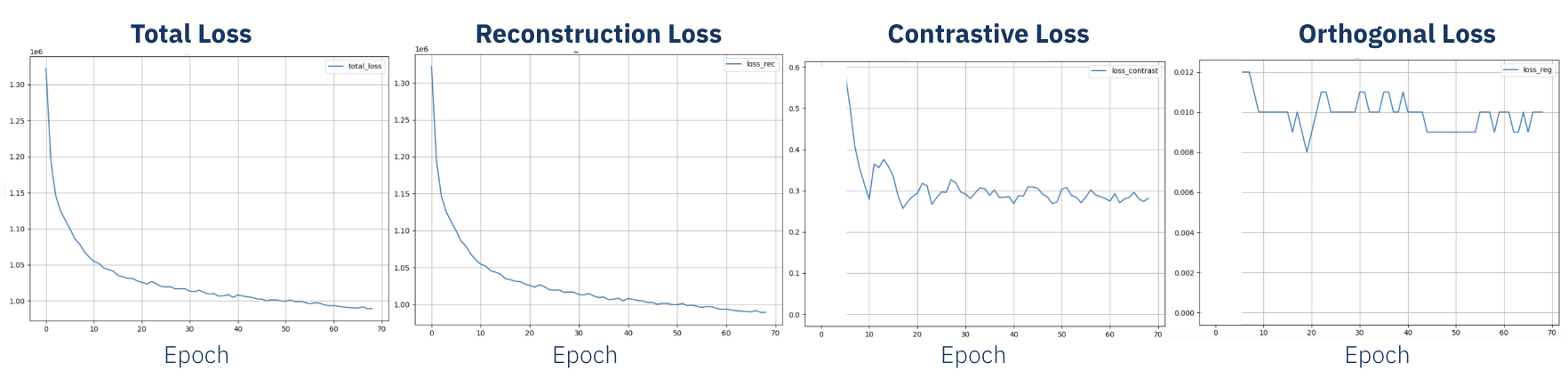}
        \caption{Representation Learning Losses}
    \end{subfigure}
    \hfill
    \begin{subfigure}[b]{1\linewidth}
        \centering
        \includegraphics[width=\linewidth]{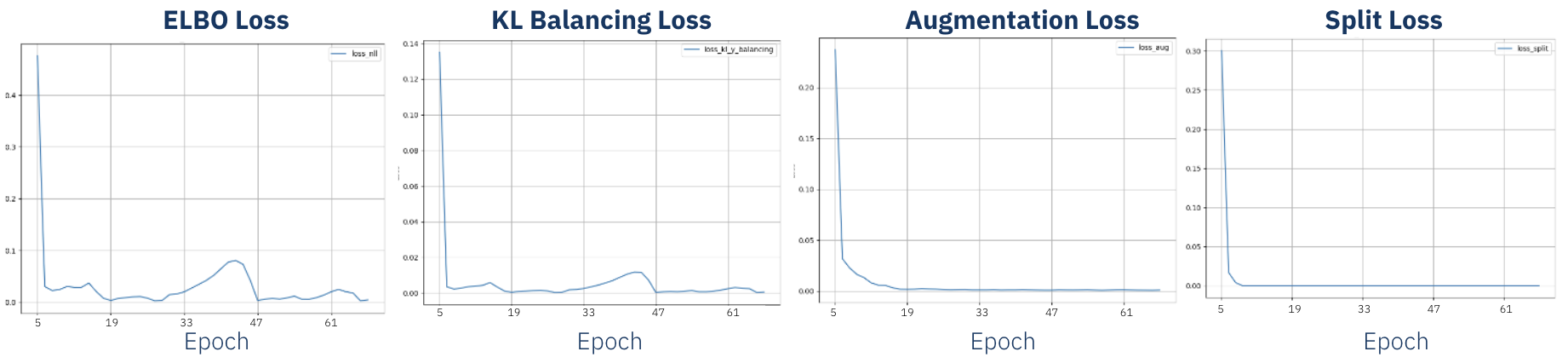}
        \caption{Dynamic Subgrouping Losses}
    \end{subfigure}
    \caption{CIFAR-10 training losses for the VAE and subgrouping models. The contrastive and orthogonal loss values are not available for the first two epochs, as joint subgroup training begins after epoch 5.}
    \label{fig:cifar_losses}
\end{figure}

\begin{figure}[htbp]
    \centering
    \begin{subfigure}[b]{1\linewidth}
        \centering
        \includegraphics[width=\linewidth]{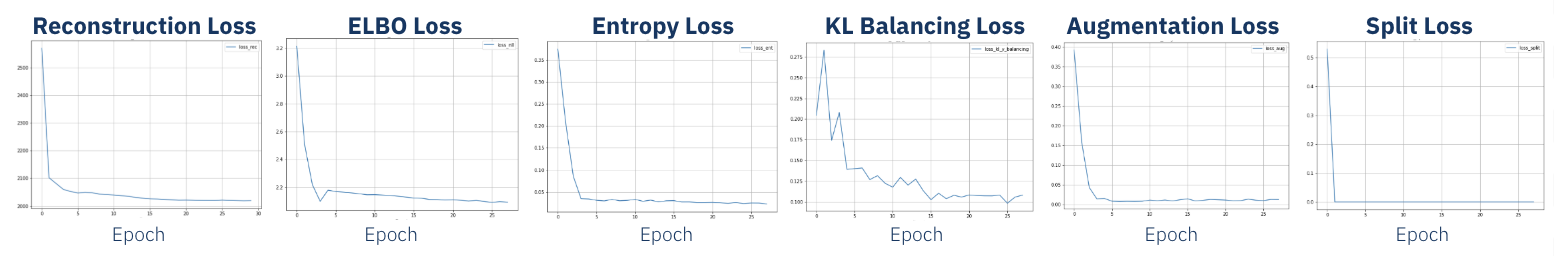}
        \caption{Blob simulated dataset training losses}
    \end{subfigure}
    \hfill
    \begin{subfigure}[b]{1\linewidth}
        \centering
        \includegraphics[width=\linewidth]{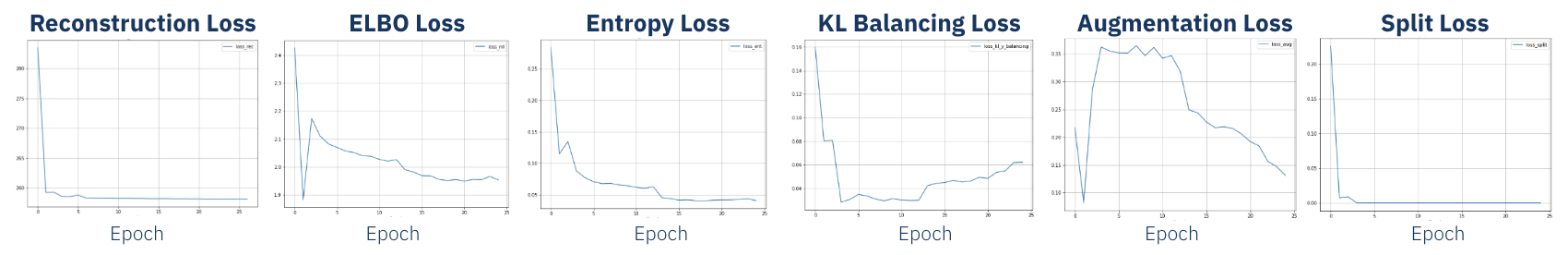}
        \caption{Moons simulated dataset training losses}
    \end{subfigure}
    \caption{Example training losses for the VAE and subgrouping models. Subgrouping begins at epoch 2 of the VAE training, so the subgrouping epochs are offset by two relative to the VAE epochs.}
    \label{fig:sim_losses}
\end{figure}

Figure~\ref{fig:cifar_losses} illustrates the training loss curves for the CIFAR-10 dataset, while Figure~\ref{fig:sim_losses} shows the curves for the Blob and Moons simulated datasets. For the Moons dataset, the augmentation loss exhibits a sudden spike followed by a gradual decrease in later epochs. Overall, all loss components demonstrate a descending and converging trend. However, occasional fluctuations, such as temporary increases followed by decreases, can occur due to the joint optimization of all loss terms. Our ablation study highlights the importance of including these components in the training process.

\subsection{Evaluation metrics: }

\paragraph{Silhouette score further explanation:} The resulting score ranges from -1 to 1: a higher score indicates better-defined clusters, a score near 0 suggests overlapping or ambiguous boundaries, and a negative score implies potential misclassification. This metric helps determine the optimal number of clusters and assess clustering quality.

% Lower Davies-Bouldin Index (DBI) value is average of all clusters {$k=[1,K]$} derived from Equation \ref{equ:dbi}. Here, $d(X_k)$ is average distance between points in cluster $k$ and the centroid of the cluster and and $D(X_k, X_{k'})$ is distance between the centroids of cluster $k$ and the nearest cluster ${k'}$. -> better clustering -> emphasize the separation between clusters
% makes no assumptions about the shape of the clusters, unlike Silhouette Score evaluation metric
% \begin{equation}
%     DBI_k = max_{k'} (\frac{d(X_k)+d(X_{k'})}{D(X_k, X_{k'})})
%     \label{equ:dbi}
% \end{equation}

\section{Additional experiments}

\subsection{OOD Images Examples}

\begin{figure}
    \centering
    \includegraphics[width=1\linewidth]{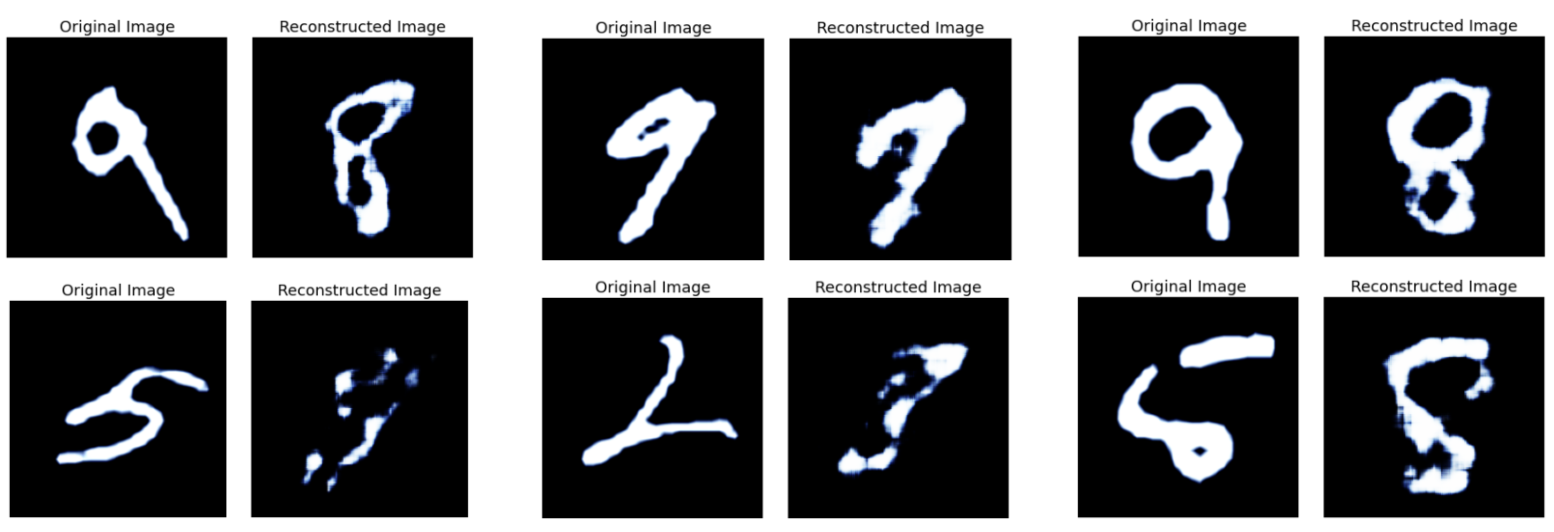}
    \caption{OOD detected original and reconstructed images.}
    \label{fig:ood_image}
\end{figure}

Figure~\ref{fig:ood_image} illustrates an example of class-OOD detection on the MNIST dataset. The top row shows examples of class 9 images that are correctly identified as OOD, as they were not present in the training data. The reconstructed images of these samples are incorrectly mapped to different digit shapes (such as 8 and 1), highlighting the importance of OOD detection in a self-supervised framework. The bottom row displays images that were also flagged as OOD but do not belong to class 9. Although the examples of them are not missed in the training data, detecting them as OOD is beneficial, as it indicates the model's limitations and points to areas where it can be improved to better represent such samples.

\begin{figure}
    \centering
    \includegraphics[width=1\linewidth]{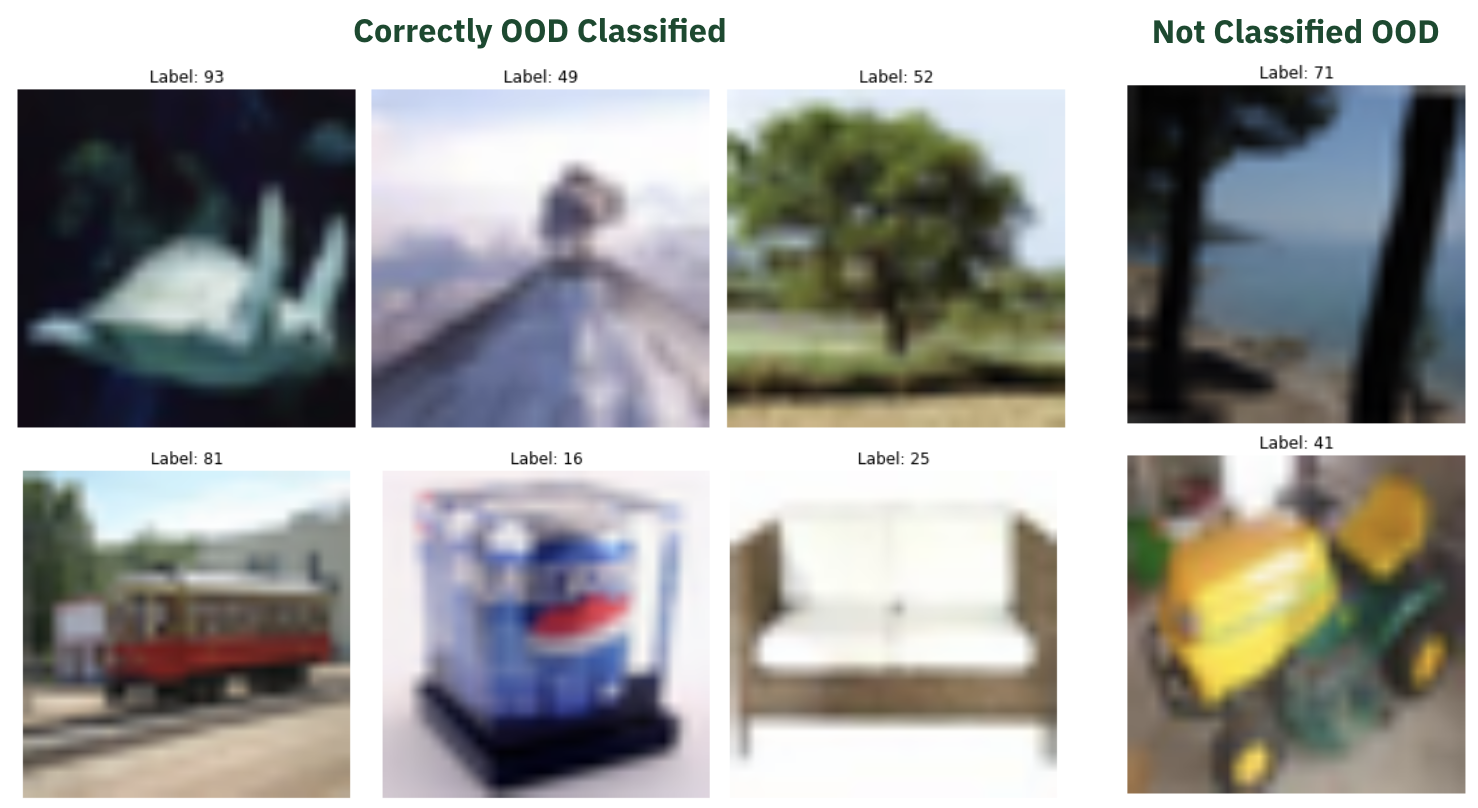}
    \caption{Near-OOD examples for CIFAR-10 and CIFAR-100: CIFAR-100 images, unseen during training, that are still correctly classified or not flagged as OOD.}
    \label{fig:far_near_ood}
\end{figure}

Examples of OOD-detected images are shown in Figure~\ref{fig:far_near_ood}, where \dynaSubVAE was trained on CIFAR-10 and tested on CIFAR-100 for near-OOD analysis. The bottom-left image is not flagged as OOD, which is reasonable since CIFAR-10 includes images of trucks, making this sample consistent with the training distribution. In contrast, the tree image in the top-left corner should ideally be classified as OOD, as such content is not represented in the CIFAR-10 dataset.

\subsection{Image Generation}
\begin{figure}[htbp]
    \centering
    \begin{subfigure}[b]{0.45\linewidth}
        \centering
        \includegraphics[width=\linewidth]{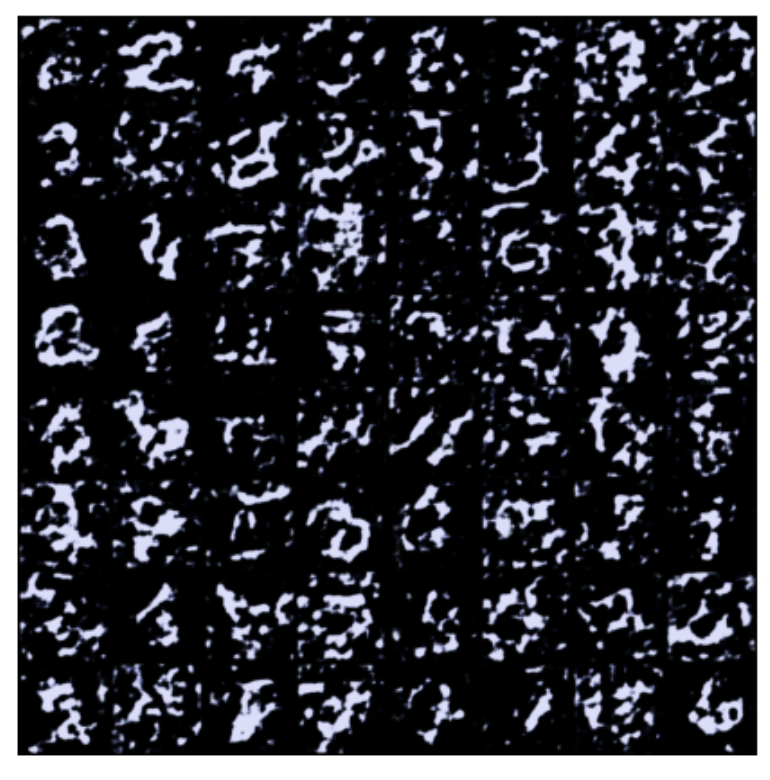}
        \caption{}
    \end{subfigure}
    \hfill
    \begin{subfigure}[b]{0.45\linewidth}
        \centering
        \includegraphics[width=\linewidth]{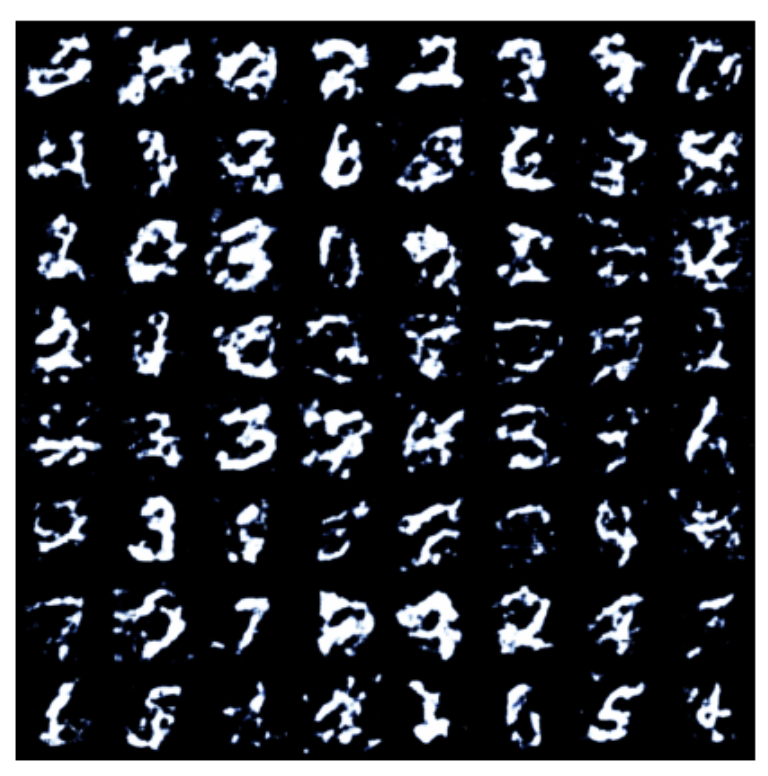}
        \caption{}
    \end{subfigure}
    \caption{KL divergence for normal distribution on $Z^{dec}$ vs $Z$.}
    \label{fig:gen_image}
\end{figure}

We evaluated OOD detection performance using two different framework designs. In the first approach, the latent variable $Z$ is derived from $H$, and $Z^{dec}$ is estimated through adaptive modulation of $Z$ based on subgroup information $C$. The KL divergence loss is applied directly to the $Z$ embeddings. In the second design, $H$ is first updated using subgroup information $C$, and then the mean ($\mu$) and log-variance ($\log(\sigma)$) are derived from the transformed $H$ to generate the final latent embeddings, to which the KL divergence loss is applied. While the first approach yields better OOD detection performance, the second approach produces more accurate image generations from random normal inputs, as illustrated in Figure~\ref{fig:gen_image}.

\begin{figure}
    \centering
    \includegraphics[width=1\linewidth]{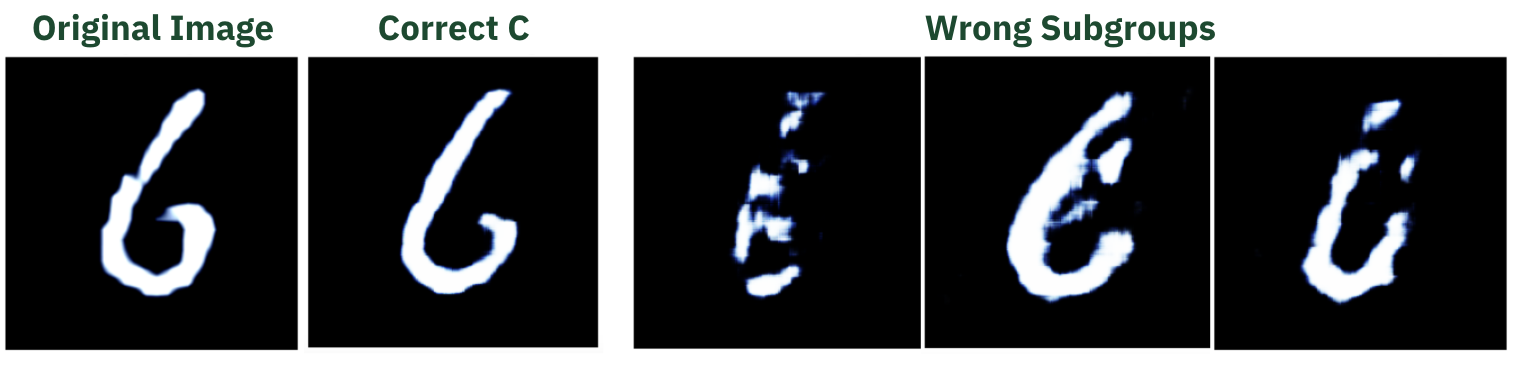}
    \caption{Impact of Changing Image Subgroup on Reconstructed Image}
    \label{fig:subgroup_impact}
\end{figure}

Figure~\ref{fig:subgroup_impact} illustrates that altering the subgroup leads to noticeable distortions in the reconstructed image, highlighting the significant impact of subgroup assignment on accurate representation. This observation motivated our definition of the regret function.

\end{document}